\documentclass[10pt,twocolumn,letterpaper]{article}

\usepackage[pagenumbers]{cvpr} % To force page numbers, e.g. for an arXiv version

\definecolor{cvprblue}{rgb}{0.21,0.49,0.74}
\usepackage[pagebackref,breaklinks,colorlinks,allcolors=cvprblue]{hyperref}
\usepackage{algorithm}
\usepackage{amsfonts} % Added for \mathbb support
\usepackage{amsmath}
\usepackage{algpseudocode}
\usepackage{multirow}
\usepackage{amsthm}
\usepackage{newfloat}
\usepackage{listings}

\theoremstyle{definition} 
\newtheorem{Definition}{Definition}
\theoremstyle{plain}  
\newtheorem{Theorem}{Theorem}
\newtheorem{Corollary}{Corollary}
\theoremstyle{plain}  
\newtheorem{TheoremS}{Theorem}
\newtheorem{CorollaryS}{Corollary}

\def\paperID{13967} % *** Enter the Paper ID here
\def\confName{CVPR}
\def\confYear{2026}

\title{Towards Reliable Evaluation of Adversarial Robustness \\ for Spiking Neural Networks}

\author{Jihang Wang$^{1,2}$ \and Dongcheng Zhao$^{1}$ \and Ruolin Chen$^{1,2}$ \and Qian Zhang$^{1,2,3,4,5}$\thanks{Corresponding author.} \and Yi Zeng$^{1,2,3,4,5*}$\\ \and
$^{1}$Brain-inspired Cognitive AI Lab, Institute of Automation, Chinese Academy of Sciences \and
$^{2}$School of Artificial Intelligence, University of Chinese Academy of Sciences\\
$^{3}$Beijing Institute of AI Safety and Governance\\
$^{4}$Beijing Key Laboratory of Safe Al and Superalignment,   
$^{5}$Center for Long-term AI\\
{\tt\small \{wangjihang2021,chenruolin2024,q.zhang,yi.zeng\}@ia.ac.cn}  \\ {\tt\small zhaodc@ion.ac.cn}
}

\begin{document}
\maketitle
\begin{abstract}
Spiking Neural Networks (SNNs) utilize spike-based activations to mimic the brain's energy-efficient information processing. However, the binary and discontinuous nature of spike activations causes vanishing gradients, making adversarial robustness evaluation via gradient descent unreliable. While improved surrogate gradient methods have been proposed, their effectiveness under strong adversarial attacks remains unclear. We propose a more reliable framework for evaluating SNN adversarial robustness. We theoretically analyze the degree of gradient vanishing in surrogate gradients and introduce the Adaptive Sharpness Surrogate Gradient (ASSG), which adaptively evolves the shape of the surrogate function according to the input distribution during attack iterations, thereby enhancing gradient accuracy while mitigating gradient vanishing. In addition, we design an adversarial attack with adaptive step size under the $L_\infty$ constraint—Stable Adaptive Projected Gradient Descent (SA-PGD), achieving faster and more stable convergence under imprecise gradients. Extensive experiments show that our approach substantially increases attack success rates across diverse adversarial training schemes, SNN architectures and neuron models, providing a more generalized and reliable evaluation of SNN adversarial robustness. The experimental results further reveal that the robustness of current SNNs has been significantly overestimated and highlighting the need for more dependable adversarial training methods. The code is released at \url{https://github.com/craree/ASSG-SNNs-Robustness-Evaluation}
\end{abstract}    
\section{Introduction}
\label{sec:intro}

Spiking Neural Networks (SNNs) have emerged as brain-inspired models that emulate the sparse, event-driven processing of biological neural systems~\cite{yin2021accurate,wu2018spatio,cai2025robust,yan2025efficient,fang2023parallel}. By transmitting information through discrete spikes instead of continuous activations, SNNs achieve high energy efficiency and rich temporal dynamics, making them attractive for neuromorphic and low-power applications~\cite{pei2019towards,roy2019towards,davies2018loihi,orchard2021efficient,ma2024darwin3}. However, the discontinuous and binary nature of spike activations hinders the application of gradient-based adversarial attacks, making reliable robustness evaluation difficult~\cite{athalye2018obfuscated}.

To overcome the non-differentiability of spike activations, surrogate gradient methods approximate the spike derivative with a smooth function~\cite{wu2018spatio,mukhoty2023direct}. While this enables backpropagation, it introduces approximation errors that distort gradients, especially under adversarial perturbations where accurate input gradients are essential~\cite{athalye2018obfuscated,athalye2018synthesizing}. As a result, gradient-based robustness evaluation in SNNs remains unreliable.

Several studies have refined surrogate gradients~\cite{bu2023rate,hao2023threaten,lun2025towards,xu2025attacking}, yet existing methods lack input-dependent designs and theoretical analysis of gradient vanishing. Moreover, most works overlook the development of constrained optimization algorithms for adversarial attacks, which are equally crucial for reliable robustness evaluation.

In this work, we propose a reliable framework for evaluating the adversarial robustness of SNNs from both the gradient approximation and attack optimization perspectives. Specifically, we first theoretically analyze the degree of gradient vanishing inherent in surrogate gradient functions of different shapes and introduce the Adaptive Sharpness Surrogate Gradient (ASSG). ASSG adaptively adjusts the sharpness of the surrogate gradient function according to the input distribution during attack iterations, enabling the surrogate gradient to closely approximate the true gradient while effectively mitigating gradient vanishing. Furthermore, we develop the Stable Adaptive Projected Gradient Descent (SA-PGD), an adversarial attack method that incorporates adaptive step-size adjustment under the $L_\infty$ constraint, enabling faster and more stable convergence even when gradients are inaccurate. Our main contributions are summarized as follows:
\begin{itemize}
\item We theoretically derive the upper bound of the gradient-vanishing degree in surrogate gradient functions during adversarial attack process. Based on this analysis, we propose the Adaptive Sharpness Surrogate Gradient (ASSG). ASSG computes an input-dependent surrogate gradient that adaptively evolves during adversarial optimization, enabling the estimation of highly accurate gradients while effectively avoiding vanishing gradients.

\item We design a new adversarial optimization algorithm with adaptive step-size, which achieves more stable and efficient convergence even when gradients are imprecise.

\item Extensive experiments across various SNN adversarial training schemes, neuron models, and architectures demonstrate that our approach substantially improves the effectiveness of adversarial attacks and generalizes well to different SNN configurations.

\item Our findings reveal that the adversarial robustness of current SNNs has been significantly overestimated. Without ASSG, SNNs may even exhibit false robustness, highlighting the urgent need to rethink how to efficiently generate reliable adversarial samples for training more genuinely robust SNN models.
\end{itemize}
\section{Related Works}
\label{sec:Related}

\subsection{Gradient Approximation Methods in SNNs}

Recent studies have shown that SNNs are vulnerable to adversarial perturbations~\cite{ding2022snn,liang2021exploring}. Surrogate gradient methods approximate the spike activation with a continuous function to enable Spatial-Temporal Backpropagation (STBP)~\cite{wu2018spatio}. The Potential-Dependent Surrogate Gradient (PDSG) method further incorporates the variance of membrane potential distributions into the surrogate computation~\cite{lun2025towards}. However, PDSG derives the surrogate function from batch-level statistics rather than individual samples, preventing adaptive modeling of sample-specific dynamics and lacking theoretical justification for its formulation.

Another line of research estimates input gradients by differentiating the SNN loss with respect to neurons’ average firing rates. Representative methods include Backward Pass Through Rate (BPTR)~\cite{ding2022snn} and Rate Gradient Approximation (RGA)~\cite{bu2023rate}. The Hybrid Adversarial attack based on Rate and Temporal information (HART)~\cite{hao2023threaten} jointly considers rate- and time-based dependencies, but its derivation is limited to a specific neuronal model and lacks theoretical validation for generalization to broader neuron types.

\subsection{Adversarial Defense Methods for SNNs}

Early studies adapted adversarial training for SNNs, such as adversarial input augmentation and min–max optimization~\cite{madry2017towards}. Later approaches, including Regularized Adversarial Training (RAT), constrain the Lipschitz constant of each layer via weight regularization to stabilize learning~\cite{ding2022snn}, while Sparsity Regularization (SR) introduces a sparse gradient penalty into the loss to enhance robustness when combined with adversarial training~\cite{liu2024enhancing}.

More recent works have strengthened SNN adversarial defense from an architectural perspective~\cite{ding2024enhancing,xu2024feel}. For instance, synaptic gating mechanisms dynamically regulate spike transmission to suppress adversarial noise propagation~\cite{ding2024enhancing}, and frequency-domain encoding schemes improve robustness by representing spiking information in more stable spectral domains~\cite{xu2024feel}.

\section{Preliminaries}
\label{sec:Preliminaries}

\subsection{Adversarial Attack and Training}

Adversarial attacks can be viewed as a constrained optimization problem in the input space of a neural network~\cite{moosavi2016deepfool,carlini2017towards,madry2017towards,croce2020reliable,dong2018boosting}. For a classification model $\boldsymbol{f}_{\boldsymbol{\theta}}:\mathbb{R}^d \rightarrow \mathbb{R}^C$, let $\boldsymbol{x} \in \mathbb{R}^d$ denote a benign input sample with the ground-truth label $y$. The objective of an untargeted adversarial attack is to find a perturbed input $\boldsymbol{x'}$ that maximizes the classification loss $\mathcal{L}$ while remaining within a small $L_p$-norm ball around the original input:
\begin{equation}
    \boldsymbol{x'} = \arg \max_{\left\lVert \boldsymbol{x'} - \boldsymbol{x} \right\rVert_p \leq \varepsilon} \mathcal{L}(\boldsymbol{f}_{\boldsymbol{\theta}}(\boldsymbol{x'}), y)
\label{1}
\end{equation}
here, $\varepsilon$ defines the perturbation bound that controls the perceptual similarity between $\boldsymbol{x}$ and $\boldsymbol{x'}$. In practice, gradient-based optimization methods such as Projected Gradient Descent (PGD) and Auto-PGD (APGD) are commonly used to approximate the solution of this problem under various norm constraints (\eg $L_\infty$, $L_2$)~\cite{madry2017towards,croce2020reliable}.

Adversarial Training (AT) is one of the most fundamental methods for improving a network’s adversarial robustness, and it has also been widely adopted in SNN-related studies~\cite{ding2022snn,ding2024robust,ding2024enhancing,xu2024feel,liu2024enhancing,mukhoty2025improving,kundu2021hire}. The objective of AT can be formulated as a min–max optimization problem:
\begin{equation}
\min _{\boldsymbol{\theta}} \mathbb{E}_{(\boldsymbol{x}, y) \sim \mathcal{D}}\left[\max _{\left\|\boldsymbol{x}^{\prime}-\boldsymbol{x}\right\|_p \leq \varepsilon} \mathcal{L}\left(\boldsymbol{f}_{\boldsymbol{\theta}}\left(\boldsymbol{x}^{\prime}\right), y\right)\right]
\label{2}
\end{equation}
where $\mathcal{D}$ denotes the data distribution. TRadeoff-inspired Adversarial DEfense via Surrogate-loss minimization (TRADES) is one of the most influential variants of AT~\cite{zhang2019theoretically,pang2022robustness}, which is also employed as a training framework for developing robust SNNs in this work.

\subsection{Spiking Neurons and Surrogate Gradients}

Each neuron in SNN integrates incoming currents into its membrane potential and generates a spike once the potential exceeds a firing threshold:
\begin{equation}
    \boldsymbol{V}^l(t+1) = F\left[ R(\boldsymbol{V}^l(t), \boldsymbol{s}^l(t)), \boldsymbol{W}^l \boldsymbol{s}^{l-1}(t+1) \right] 
    \label{3}
\end{equation}
\begin{equation}
    \boldsymbol{s}^l(t+1) = H(\boldsymbol{V}^l(t+1) - V_{\text{th}})
    \label{4}
\end{equation}
where $\boldsymbol{V}^l(t)$, $\boldsymbol{s}^l(t)$, and $\boldsymbol{W}^l$ denote the membrane potential, spike output, and synaptic weight matrix in layer $l$ at time step $t$, respectively. $H(x)$ is the Heaviside step function. $F(\cdot)$ and $R(\cdot)$ represent the membrane potential update and reset mechanisms, respectively. Different neuron models adopt different definitions for these two functions.

In this paper, we primarily employ the Leaky Integrate-and-Fire (LIF) neuron model:
\begin{equation}
    \boldsymbol{V}^l(t+1) = \lambda R\!\left(\boldsymbol{V}^l(t), \boldsymbol{s}^l(t)\right) + (1 - \lambda)\, \boldsymbol{W}^l \boldsymbol{s}^{l-1}(t+1),
    \label{5}
\end{equation}
\begin{equation}
R\!\left(\boldsymbol{V}^l(t), \boldsymbol{s}^l(t)\right)= 
\begin{cases}
0, & \boldsymbol{s}^l(t) = 1,\\[3pt]
\boldsymbol{V}^l(t), & \boldsymbol{s}^l(t) = 0.
\end{cases}
\label{6}
\end{equation}
Here, $\lambda \in (0,1)$ is the membrane potential decay coefficient. The reset mechanism adopts a hard reset. 

To evaluate the generality of the proposed attack, we employ several neuron models in addition to the standard LIF neuron, including its variant LIF-2, the Integrate-and-Fire (IF) neuron, and the Parallel Spiking Neuron (PSN)~\cite{fang2023parallel}. Detailed configurations of these models are provided in the Appendix \cref{S2}.

Regardless of the neuron model, backpropagation in SNNs requires approximating $\frac{dH}{dx}$ through surrogate gradients. For example, the arctangent (Atan) surrogate gradient function is defined as:
\begin{equation}
\frac{dH}{dx} \approx g(x) = \frac{\alpha}{2 \left[1 + (\frac{\pi}{2} \alpha x)^2 \right]},
\label{7}
\end{equation}
where $\alpha$ controls the sharpness of the surrogate gradient. In the Appendix \cref{S1}, we provide additional examples of surrogate gradient functions.

\section{Methods}
\label{sec:Method}
\begin{figure*}[h]
  \centering
   \includegraphics[width=0.96\linewidth]{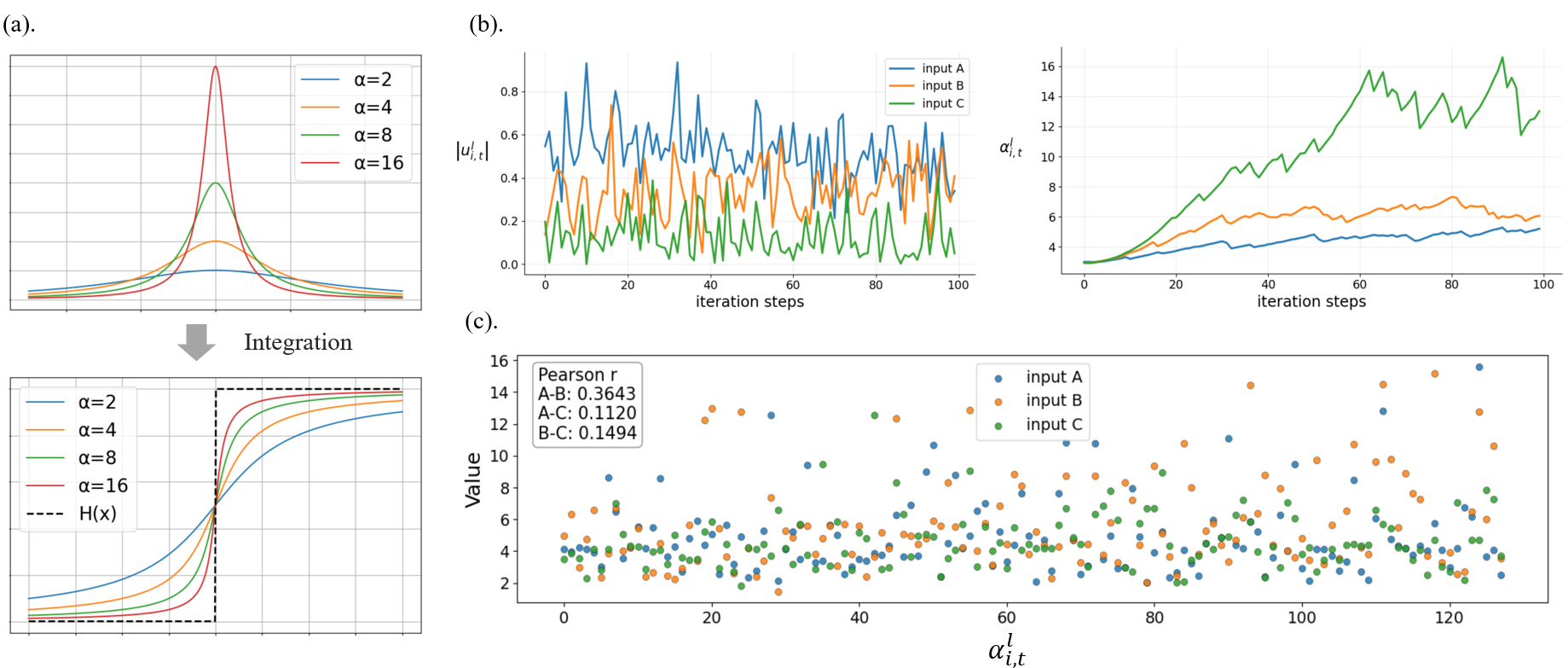}
   \caption{(a). Curves of the surrogate gradient function $g(x)$ and its integral under different sharpness parameters $\alpha $. (b). Heterogeneous evolution of $\left|u_{i,t}^l\right|$ and $\alpha_{i,t}^l$ across different inputs at a fixed $(i,t,l)$ during adversarial optimization. (c) spatial-temporal heterogeneity of a group of $\alpha_{i,t}^l$ across different inputs. The Pearson correlation coefficients of the input-dependent $\alpha_{i,t}^l$ are also shown in the figure.}
   \label{assg}
\end{figure*}
\subsection{Theoretical Analysis on Gradient Vanishing}

Surrogate gradient is employed to approximate a non-zero and smooth derivative of $H(x)$ for training or adversarial attack. We define the surrogate gradient as the following general class of functions:
\begin{Definition}
A function $g:\mathbb{R} \rightarrow \mathbb{R}_{\geq 0}$ is a surrogate function if it is even, non-decreasing on the interval $(-\infty,0)$ and $\int_{-\infty}^{\infty} g(x) \,dx=1$
\label{d1}
\end{Definition}
Unlike the definition given by Mukhoty et al.~\cite{mukhoty2023direct}, we impose the normalization constraint $\int_{-\infty}^{\infty} g(x) \,dx=1$. This property guarantees that the integral of $g(x)$ behaves as an approximation to $H(x)$. As shown in \cref{assg}(a), $g(x)$ can take either a narrow and sharp or wide and flat shape; a sharper function makes its integral closer to $H(x)$, implying a more accurate gradient approximation.

However, a sharper $g(x)$ tends to approach 0 more rapidly as $\left\lvert x \right\rvert$ increases, which can lead to gradient vanishing.
Our motivation is to assign $g(x)$ an appropriate sharpness that closely approximates $H(x)$ while preventing excessive gradient vanishing. To this end, we quantitatively measure the gradient vanishing of $g(x)$ as follows:
\begin{equation}
G(x) = \int_{-\left\lvert x \right\rvert}^{\left\lvert x \right\rvert} g(t) \, dt,
\label{8}
\end{equation}
where $G(x)$ denotes the gradient-vanishing degree of $g(x)$ at position $x$. Without loss of generality, we define the domain of $G(x)$ as $[0, +\infty)$. It can be observed that $G(x)$ is a monotonically non-decreasing function whose range lies within $[0,1]$. A larger value of $G(x)$ indicates a higher degree of gradient vanishing.

$G(x)$ serves as a suitable quantitative indicator for measuring the gradient vanishing in surrogate functions. As the surrogate function $g(x)$ becomes sharper, $G(x)$ rises toward 1 more rapidly. For the exact derivative of the Heaviside function $H(x)$, we have $G(x) = 1$ on $\forall x > 0$, which implies that the gradient vanishes everywhere.

We can derive the following theorem about $G(x)$:
\begin{Theorem}
For any surrogate function $g(x)$, the gradient-vanishing degree function $G(x) = \int_{-x}^{x} g(t) \, dt,x \geq 0$ is a concave function on $[0,+\infty)$
\label{t1}
\end{Theorem}
We aim to control the gradient-vanishing degree by adjusting the sharpness of the surrogate function. Without loss of generality, we express the surrogate function in the form $\alpha k(\alpha x)$, whose primitive function can be written as $\int _{-\infty}^x \alpha k(\alpha x)dx=K(\alpha x)$. The parameter $\alpha$ in this surrogate function can be used to control its sharpness. As shown in \cref{assg}(a), a larger $\alpha$ leads to a sharper surrogate function, and its primitive function becomes closer to $H(x)$, indicating more accurate gradient estimation.

From Theorem \ref{t1}, we can select an appropriate sharpness parameter $\alpha$ such that the expected gradient-vanishing degree is upper-bounded by a prescribed constant:
\begin{Corollary}
    \label{c1}
    Let $g(x)=\alpha k(\alpha x)$ a surrogate function, and the corresponding  gradient-vanishing degree function $\int_{-x}^{x} \alpha k(\alpha t) \, dt =  \int_{-\alpha x}^{\alpha x} k(\alpha t) \, d\alpha t = G(\alpha x), x \geq 0$. Assume that $x$ follows a probability distribution $p(x)$ with finite expectation $\mathbb{E}[x] < +\infty$. Then, for any constant $A \in [0,1]$, the expected gradient-vanishing degree satisfies 
    \begin{equation}
    \mathbb{E}_{x\!\sim\!p(x)}[G(\alpha x)] \le A
    \;\text{if}\;
    \alpha \le \frac{G^{-1}(A)}{\mathbb{E}[x]}
    \label{9}
    \end{equation}
\end{Corollary}
The detailed proof required for this section is provided in the Appendix \cref{S1}.

\subsection{Adaptive Sharpness Surrogate Gradient}

Based on the theoretical analysis above, ASSG adaptively computes the sharpness parameter across spatial-temporal dimension of every input during adversarial attacks.

For a given input sample $\boldsymbol{x}$, an adversarial attack iteratively searches for adversarial examples, forming an optimization trajectory $\{\boldsymbol{x_1}, \boldsymbol{x_2},...,\boldsymbol{x_k},...\}$. $\boldsymbol{x_k}$ is transformed into the input to $H$ at a specific spatial-temporal dimension $(i,t,l)$: $u_{i,t}^l(\boldsymbol{x_k})=V_i^l(t+1) - V_{\text{th}}$ ($i$ denotes the neuron index). Let $p(\left\lvert u_{i,t}^l \right\rvert)$ denote the probability distribution of $\left\lvert u_{i,t}^l \right\rvert$ during the optimization proces. According to Corollary \ref{c1}, for surrogate functions of the form $\alpha k(\alpha x)$, we can compute $\alpha_{i,t}^l$ to constrain the expected upper bound of gradient vanishing throughout the optimization trajectory:
\begin{equation}
    \alpha_{i,t}^l = \frac{\omega }{\mathbb{E}[\left\lvert u_{i,t}^l \right\rvert]}
    \label{10}
\end{equation}
where $\omega =G^{-1}(A)$. We replace the inequality in \cref{9} with an equality, as we prefer $\alpha_{i,t}^l$ to take the maximum admissible value under the constraint $\mathbb{E}_{\left\lvert u_{i,t}^l \right\rvert \sim p(\left\lvert u_{i,t}^l \right\rvert)}\!\left[ G(\left\lvert u_{i,t}^l \right\rvert) \right] \le A$, ensuring that the surrogate gradient approximates $\frac{dH}{dx}$ as closely as possible. $\omega$ is tunable to adjust the upper bound of gradient-vanishing degree. To obtain the optimal gradient accuracy-vanishing trade-off for attack, we select the best $\omega$ via ternary search. 

For different surrogate functions, the valid range of $\omega$ may vary. However, we uniformly set its range to $[0, +\infty)$. In the Appendix \cref{S1}, we provide a detailed explanation of this range and give examples of the specific computation of $\omega$ for various surrogate functions.

Next, we estimate $\mathbb{E}[\left\lvert u_{i,t}^l \right\rvert]$ during adversarial optimization process. ASSG adaptively estimates $\mathbb{E}[\left\lvert u_{i,t}^l \right\rvert]$ by maintaining an exponential moving average (EMA) of $\left\lvert u_{i,t}^l \right\rvert$, which is then used to dynamically update the sharpness parameter $\alpha_{i,t}^l$ throughout the attack iterations:
\begin{equation}
M_{i,t}^l(\boldsymbol{x_{1:k}}) = \beta\,M_{i,t}^l(\boldsymbol{x_{1:k-1}}) + (1 - \beta)\,|u_{i,t}^l(\boldsymbol{x_k})|
\label{11}
\end{equation}
\begin{equation}
    \alpha_{i,t}^l(\boldsymbol{x_{1:k}}) = \frac{\omega }{M_{i,t}^l(\boldsymbol{x_{1:k}})}.
    \label{12}
\end{equation}
where $\beta \in (0,1)$ is the momentum coefficient, $\boldsymbol{x_{1:k}}$ denotes the set of points up to the $k$-th iteration.

ASSG further incorporates an exponential moving average of the absolute deviation of $\left\lvert u_{i,t}^l \right\rvert$ as a relaxation term on top of the EMA estimate:
\begin{equation}
M_{i,t}^l(\boldsymbol{x_{1:k}}) = \beta_1\,M_{i,t}^l(\boldsymbol{x_{1:k-1}}) + (1 - \beta_1)\,|u_{i,t}^l(\boldsymbol{x_{k}})|
\label{13}
\end{equation}
\begin{equation}
\begin{aligned}
D_{i,t}^l(\boldsymbol{x_{1:k}}) 
&= \beta_2\,D_{i,t}^l(\boldsymbol{x_{1:k-1}}) \\
&\quad + (1 - \beta_2)\,\bigl|u_{i,t}^l(\boldsymbol{x_k}) - M_{i,t}^l(\boldsymbol{x_{1:k}})\bigr|
\end{aligned}
\label{eq:14}
\end{equation}
\begin{equation}
    \alpha_{i,t}^l(\boldsymbol{x_{1:k}}) = \frac{\omega }{ M_{i,t}^l(\boldsymbol{x_{1:k}}) + \gamma D_{i,t}^l(\boldsymbol{x_{1:k}})}.
    \label{15}
\end{equation}
where $\beta_1, \beta_2 \in (0,1)$ are the momentum coefficients, $\gamma$ is the relaxation coefficient. The relaxation term is introduced to slightly reduce $\alpha_{i,t}^l$ by incorporating the second-order moment information of the $\left\lvert u_{i,t}^l \right\rvert$ distribution, thereby ensuring that the gradient-vanishing degree remains within the theoretical upper bound.

$M_{i,t}^l$ and $D_{i,t}^l$ are initialized to $M_0$ and $D_0$. \cref{13}-\cref{15} together constitute the adaptive update rule of ASSG, which allows each neuron to self-adjust its surrogate sharpness.

\begin{figure}[t]
  \centering
   \includegraphics[width=0.96\linewidth]{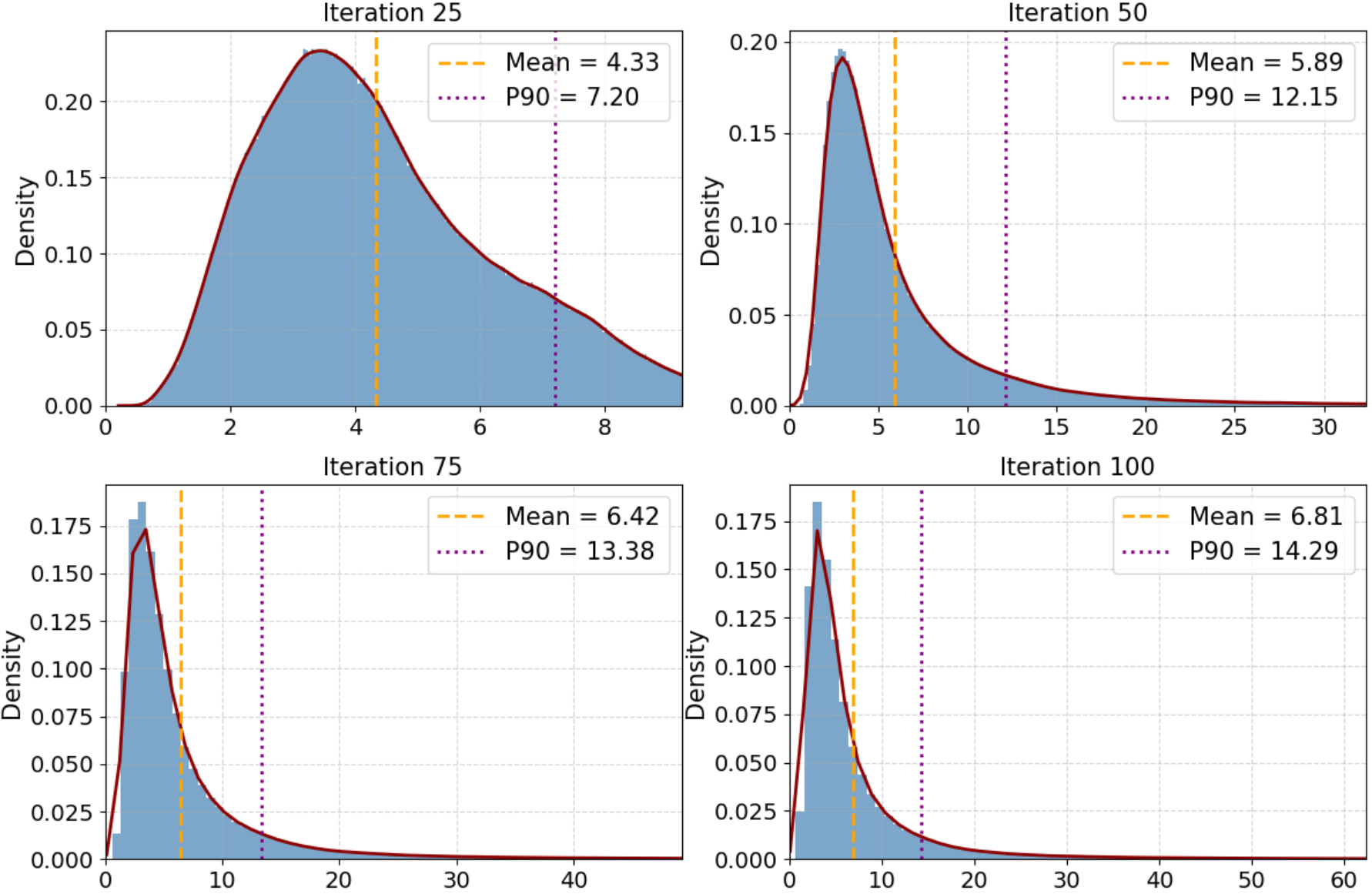}
   \caption{The distribution of the sharpness parameter $\alpha_{i,t}^l$. The vertical lines denote the mean and the 90th percentile (P90).}
   \label{alpha}
\end{figure}

In \cref{assg}(b), We visualize, at a fixed spatial-temporal location $(i,t,l)$, the dynamics of $\left\lvert u_{i,t}^l \right\rvert$ for different inputs during the adversarial attack process, together with the input-adaptive $\alpha_{i,t}^l$ computed by ASSG. It can be observed that the closer the distribution of $\left|u_{i, t}^l\right|$ is to 0, the larger the $\alpha_{i, t}^l$ evolved by ASSG.

In \cref{assg}(c), we visualize the distribution differences of $\alpha_{i,t}^l$ at a group of spatial-temporal locations $(i,t,l)$ for different inputs at the 75th iteration of the adversarial attack. The figure shows that the pairwise Pearson correlation coefficients among the three groups of $\alpha_{i,t}^l$ are all very low, further demonstrating that different inputs lead to highly heterogeneous spatial-temporal distributions of $\alpha_{i,t}^l$. This indicates that ASSG possesses strong input-adaptive capability across both spatial and temporal dimensions.

In \cref{alpha}, we visualize the distribution of the sharpness parameter $\alpha_{i,t}^l$ across different spatial-temporal dimensions in ASSG and its evolution along the optimization trajectory. As shown, $\alpha_{i,t}^l$ gradually develops into a more dispersed unimodal, long-tailed distribution as the iterations proceed, indicating that both sharp and smooth surrogate gradient shapes coexist within the SNN. 

\begin{figure}[t]
  \centering
   \includegraphics[width=0.96\linewidth]{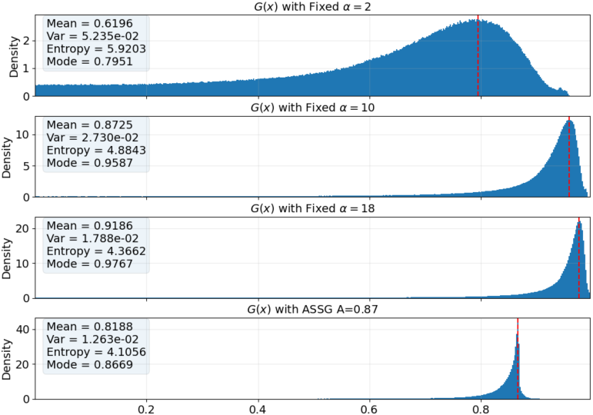}
   \caption{The distribution of the degree of gradient vanishing $G(x)$ in the SNN using the Atan surrogate gradient with different fixed $\alpha$ and with ASSG ($A = 0.87$)}
   \label{GVD}
\end{figure}

In \cref{GVD}, we perform backpropagation in the SNN using the Atan surrogate gradient with different fixed $\alpha$ values, as well as with ASSG that adaptively computes $\alpha_{i, t}^l$ using $A = 0.87$, and visualize the distribution of gradient-vanishing degree $G(x)$ for neurons across all spatial-temporal dimensions. When fixed $\alpha$ is used, the distribution of $G(x)$ in the SNN is more dispersed (with higher entropy). As $\alpha$ increases, the distribution shifts toward 1. In contrast, ASSG effectively concentrates the gradient vanishing degrees in the SNN around the theoretically upper bound $A$ (with the distribution peak at 0.8669).

In summary, \cref{assg}(b)(c) demonstrates the input-dependent property of ASSG, while \cref{alpha} shows that, for each input, ASSG evolves a highly spatial-temporally heterogeneous distribution of $\alpha_{i, t}^l$. Furthermore, ASSG enables the surrogate gradients of every neuron to be as accurate as possible (as reflected by the long-tailed distribution in \cref{alpha}), while constraining the gradient-vanishing degree in the SNN to be tightly concentrated within a specified upper bound (\cref{GVD}). Consequently, by adjusting this upper bound, we can allow ASSG to perform backpropagation under different gradient accuracy-vanishing trade-offs, thereby achieving stronger attack performance.

\subsection{Stable Adaptive Projected Gradient Descent}

We find that traditional optimization algorithms, such as PGD and APGD, fail to fully converge even when combined with ASSG for attacking robust SNNs—even after hundreds of iterations, as shown in \cref{sapgd-abla}. This suggests that the input space of SNNs is highly non-smooth and that the gradients remain inaccurate. These observations underscore the need for a more effective adversarial attack optimization algorithm specifically tailored to SNNs—an aspect largely overlooked in prior research.

SA-PGD incorporates both adaptive step size and momentum into adversarial attack optimization, while also considering the $L_\infty$-norm constraint.

At each attack iteration in SA-PGD, we first compute the $L_1$-normalized momentum and the degree of gradient oscillation as follows:
\begin{equation}
\boldsymbol{m}_{k} = \beta_m \boldsymbol{m}_{k-1} + \frac{\nabla_{\boldsymbol{x}} \mathcal{L}\left(\boldsymbol{f}_{\boldsymbol{\theta}}\left(\boldsymbol{x}_{k}\right), y\right))}
{\|\nabla_{\boldsymbol{x}} \mathcal{L}\left(\boldsymbol{f}_{\boldsymbol{\theta}}\left(\boldsymbol{x}_{k}\right), y\right)\|_1},
\label{16}
\end{equation}
\begin{equation}
\boldsymbol{v}_{k} = \beta_v \boldsymbol{v}_{k-1} + \frac{\nabla_{\boldsymbol{x}} \mathcal{L}\left(\boldsymbol{f}_{\boldsymbol{\theta}}\left(\boldsymbol{x}_{k}\right), y\right)^2}
{\|\nabla_{\boldsymbol{x}} \mathcal{L}\left(\boldsymbol{f}_{\boldsymbol{\theta}}\left(\boldsymbol{x}_{k}\right), y\right)^2\|_1},
\label{17}
\end{equation}
where $\beta_m, \beta_v$ are momentum coefficients. Next we compute the adaptive step size for momentum update based on $\boldsymbol{v}_{k}$, and apply clipping under the $L_\infty$-norm constraint to suppress excessively large updates:
\begin{equation}
\boldsymbol{t}_k =
\mathrm{clip}\!\left(
\frac{\boldsymbol{m}_k}{\sqrt{\boldsymbol{v}_k} + \xi}
\cdot \eta_k,\;
-\eta_k,\; \eta_k
\right),
\label{18}
\end{equation}
\begin{equation}
\boldsymbol{x}_{k+1} = \Pi_{\varepsilon}^{\infty}\!\left(\boldsymbol{x}_{k} + \boldsymbol{t}_{k} \right)
\label{19}
\end{equation}
where $\eta_k$ is the learning rate, and we adopt the same learning rate adjustment strategy as in APGD. $\xi$ is a small constant for numerical stability. The clipping operator in \cref{18} limits each element of the update within $[-\eta_k,\eta_k]$, which ensures that no single dimension dominates the adversarial update, maintaining stable and bounded optimization under the $L_\infty$-norm constraint. $\Pi_{\varepsilon}^{\infty}$ projects the updated input back into the feasible region defined by the $L_\infty$-norm constraint with a perturbation radius of $\varepsilon$.

\section{Experiments}
\label{sec:Experiments}

\begin{table*}[t]
  \centering
  \caption{Comparison with the state-of-the-art works. The best results are in bold.}
  \label{tab1}
  \begin{tabular}{c|c|cccccccc}
  \hline
 Datasets & Methods & Clean & Attack & BPTR~\cite{ding2022snn} & RGA~\cite{bu2023rate} & PDSG~\cite{lun2025towards} & HART~\cite{hao2023threaten} & STBP~\cite{wu2018spatio} & \textbf{ASSG} \\
   \hline
\multirow{10}{*}{\shortstack{CIFAR \\ -10}} & \multirow{2}{*}{AT} & \multirow{2}{*}{78.69} & APGD & 66.45 & 63.94 & 69.31 & 77.22 & 75.38 & \textbf{84.06} \\
                                                                                     & & & SA-PGD & 66.85 & 65.08 & 68.76 & 79.49 & 75.11 & \textbf{88.44} \\
                            \cline{2-10}
    & \multirow{2}{*}{RAT} & \multirow{2}{*}{80.03} & APGD & 61.81 & 61.05 & 67.38 & 73.15 & 70.81 & \textbf{81.49} \\
                                              & & & SA-PGD & 62.67 & 62.45 & 66.72 & 75.66 & 70.99 & \textbf{86.44} \\
                           \cline{2-10}
    & \multirow{2}{*}{AT+SR} & \multirow{2}{*}{77.93} & APGD & 54.51 & 52.62 & 56.20 & 65.77 & 60.82 & \textbf{73.76} \\
                                                & & & SA-PGD & 54.63 & 52.65 & 55.43 & 67.71 & 60.37 & \textbf{77.00} \\
                            \cline{2-10}
    & \multirow{2}{*}{TRADES} & \multirow{2}{*}{81.06} & APGD & 59.75 & 56.97 & 59.15 & 70.63 & 66.27 & \textbf{75.91} \\
                                                 & & & SA-PGD & 60.00 & 57.64 & 58.33 & 72.47 & 66.24 & \textbf{79.56} \\
                            \cline{2-10}
  \hline
\multirow{10}{*}{\shortstack{CIFAR \\ -100}} & \multirow{2}{*}{AT} & \multirow{2}{*}{51.90} & APGD & 82.83 & 82.50 & 84.21 & 89.88 & 88.22 & \textbf{91.60} \\
                                                                                      & & & SA-PGD & 82.99 & 83.07 & 84.00 & 90.80 & 88.26 & \textbf{93.19} \\
                            \cline{2-10}
    & \multirow{2}{*}{RAT} & \multirow{2}{*}{55.38} & APGD & 76.74 & 77.28 & 79.68 & 84.77 & 83.33 & \textbf{87.73} \\
                                              & & & SA-PGD & 76.94 & 78.35 & 79.43 & 86.21 & 83.26 & \textbf{89.64} \\
                           \cline{2-10}
    & \multirow{2}{*}{AT+SR} & \multirow{2}{*}{52.11} & APGD & 76.86 & 77.80 & 78.88 & 85.09 & 82.84 & \textbf{86.18} \\
                                                & & & SA-PGD & 76.90 & 78.19 & 78.37 & 85.76 & 82.73 & \textbf{87.68} \\
                            \cline{2-10}
    & \multirow{2}{*}{TRADES} & \multirow{2}{*}{57.08} & APGD & 78.89 & 78.99 & 79.17 & 86.07 & 84.27 & \textbf{87.57} \\
                                                 & & & SA-PGD & 79.06 & 79.39 & 78.88 & 87.00 & 84.15 & \textbf{89.22} \\
                            \cline{2-10}
    \hline           
    \multirow{10}{*}{\shortstack{CIFAR10 \\ -DVS}} & \multirow{2}{*}{AT} & \multirow{2}{*}{79.70} & APGD & 33.00 & 32.10 & 33.40 & 35.40 & 36.10 & \textbf{47.00}\\
                                                                                            & & & SA-PGD & 32.00 & 31.90 & 33.10 & 34.80 & 35.40 & \textbf{49.10} \\
                            \cline{2-10}
    & \multirow{2}{*}{RAT} & \multirow{2}{*}{82.50} & APGD & 31.30 & 30.20 & 33.20 & 34.00 & 35.20 & \textbf{42.50} \\
                                              & & & SA-PGD & 30.60 & 29.20 & 31.80 & 33.20 & 34.30 & \textbf{44.60} \\
                           \cline{2-10}
    & \multirow{2}{*}{AT+SR} & \multirow{2}{*}{79.40} & APGD & 31.90 & 31.40 & 32.50 & 34.50 & 36.30 & \textbf{42.50} \\
                                                & & & SA-PGD & 31.20 & 31.00 & 31.90 & 34.60 & 35.60 & \textbf{43.60} \\
                            \cline{2-10}
    & \multirow{2}{*}{TRADES} & \multirow{2}{*}{76.60} & APGD & 32.20 & 30.70 & 32.70 & 34.00 & 34.20 & \textbf{36.20} \\
                                                 & & & SA-PGD & 31.30 & 28.80 & 32.00 & 33.60 & 33.90 & \textbf{36.80} \\
                            \cline{2-10}
    \hline     
  \end{tabular}
\end{table*}

\begin{table}[h]
  \centering
  \caption{ASR on different surrogate gradient functions.}
  \label{ASSG-abla}
  \begin{tabular}{cccccc}
  \hline
 Method & Atan & Rect & Tri & Sig & Gauss \\
   \hline
     \multicolumn{6}{c}{CIFAR-10} \\
  \hline
    fixed $\alpha$ & 84.04 & 77.90 & 81.07 & 81.89 & 81.29 \\
    \textbf{ASSG} & \textbf{88.44} & \textbf{86.82} & \textbf{89.53} & \textbf{89.30} & \textbf{89.46} \\
  \hline
    \multicolumn{6}{c}{CIFAR-100} \\
  \hline
    fixed $\alpha$ & 91.40 & 90.02 & 91.08 & 91.22 & 91.16 \\
    \textbf{ASSG} & \textbf{93.19} & \textbf{92.31} & \textbf{93.41} & \textbf{93.35} & \textbf{93.42} \\
    \hline     
  \end{tabular}
\end{table}

\subsection{Experimental Setup}

We conduct experiments on both static datasets (CIFAR-10 and CIFAR-100~\cite{krizhevsky2009learning}) and a dynamic dataset (CIFAR10-DVS~\cite{li2017cifar10}). For each dataset, we employ multiple adversarial training methods to obtain robust models and evaluate the attack success rates (ASR) of the proposed approach. The surrogate function in ASSG is set to the Atan function in \cref{7} by default, and the optimal $\omega$ is selected using ternary search.
% Unlike previous studies~\cite{bu2023rate,hao2023threaten,lun2025towards}, we do not consider non-robust SNNs, as these models often achieve 100\% ASR when using APGD with large iteration steps and basic gradient approximation methods.

We train robust SNN models using AT, RAT, AT combined with SR (AT+SR), and TRADES. For robust training, we generate adversarial samples with PGD using 5 iterations and a perturbation magnitude $\varepsilon=8/255$, which is a common practice in current SNN research. The default neuron model in our SNN is the LIF neuron, as described in \cref{5}-\cref{6}, with a time step of 4 for CIFAR-10/100 and 8 for CIFAR10-DVS. For CIFAR-10/100, the network structure is SEWResNet19~\cite{fang2021deep}, while for CIFAR10-DVS, we use VGG9~\cite{duan2022temporal}. The default input encoding is direct encoding. The detailed experimental setup and parameter configuration are provided in the Appendix \cref{S3}.

\subsection{Comparison with State-of-the-art Works}

For gradient approximation methods, we compare ASSG with basic STBP (which uses surrogate gradients during training for adversarial attacks)~\cite{wu2018spatio} and with recent methods BPTR~\cite{ding2022snn}, RGA~\cite{bu2023rate}, HART~\cite{hao2023threaten}, and PDSG~\cite{lun2025towards}. Implementations follow the original papers. For attack methods, APGD serves as the main baseline, with a default of 100 attack iterations and $\varepsilon=8/255$. 

\Cref{tab1} shows the ASR for different combinations of surrogate gradient approximation methods and attack optimization algorithms. A higher ASR indicates a more accurate evaluation of adversarial robustness. ASSG consistently achieves the highest ASR across various datasets and training methods, particularly on CIFAR-10 and CIFAR10-DVS, where ASSG outperforms other gradient approximation methods by a significant margin (near 10\% in 7 cases). When combined with SA-PGD, ASSG further improves ASR over APGD. However, SA-PGD shows little benefit when paired with low-quality gradient approximations (BPTR, RGA, PDSG, STBP), indicating that the optimization algorithm and the gradient approximation method mutually influence the effectiveness of adversarial attacks.

\begin{figure}[t]
  \centering
   \includegraphics[width=0.96\linewidth]{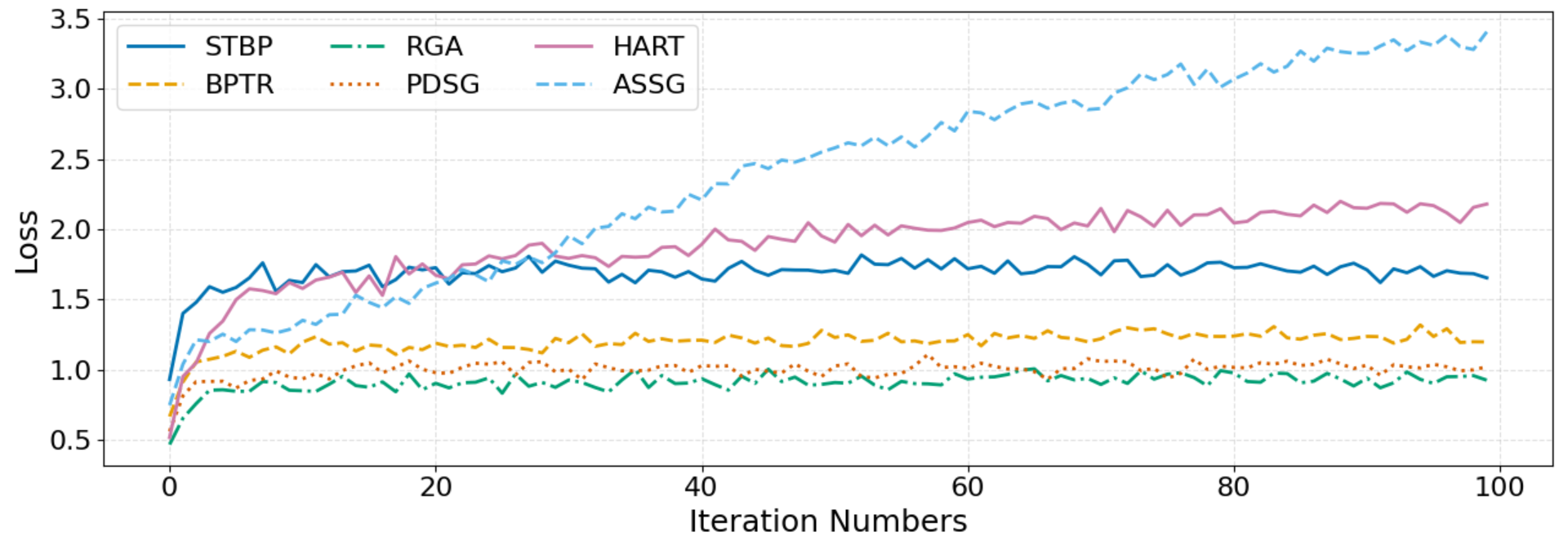}
   \caption{The loss curves of SA-PGD for different gradient approximation methods.}
   \label{loss}
\end{figure}

Furthermore, we plot the loss curves of SA-PGD on CIFAR-10 for different gradient approximation methods in \cref{loss}. The loss curves of STBP and HART are higher than that of ASSG in the early stages but are later surpassed by ASSG as the iterations proceed. As the number of iterations increases, the loss curves of all gradient approximation methods except ASSG tend to converge, while ASSG continues to exhibit an upward trend, demonstrating the effectiveness of ASSG in adaptively evolving the distribution of $\alpha_{i, t}^l$ throughout the attack.

In our configuration, the ASR of RGA and PDSG methods are even lower than that of basic STBP, which contradicts the conclusions of their original papers~\cite{bu2023rate,lun2025towards}. However, our results are more reliable, as previous works~\cite{bu2023rate,hao2023threaten,lun2025towards} only compared ASR using PGD with 10 iterations. The higher ASR reported in those studies may simply reflect a faster loss increase during the early attack iterations.

\begin{table*}[h]
  \centering
  \caption{ASR on different neuron models, time steps $T$ and encoder.}
  \label{tab3}
  \begin{tabular}{c|ccc|cccccccc}
  \hline
 Datasets & Neuron & $T$ & Encoder & Clean & Attack & BPTR & RGA & PDSG & HART & STBP & \textbf{ASSG} \\
   \hline
\multirow{10}{*}{\shortstack{CIFAR \\ -10}} & \multirow{2}{*}{LIF-2} & \multirow{2}{*}{4} & \multirow{2}{*}{Direct} & \multirow{2}{*}{79.68}
                                & APGD & 63.01 & 63.52 & 65.82 & 71.99 & 71.93 & \textbf{78.30} \\
                      & & & & & SA-PGD & 63.36 & 63.11 & 65.27 & 73.50 & 73.15 & \textbf{82.40} \\
                            \cline{2-12}
    & \multirow{2}{*}{IF} & \multirow{2}{*}{4} & \multirow{2}{*}{Direct} & \multirow{2}{*}{77.21}
                                & APGD & 67.16 & 67.25 & 68.92 & 64.14 & 75.38 & \textbf{86.51} \\
                      & & & & & SA-PGD & 66.83 & 67.04 & 68.13 & 65.22 & 76.54 & \textbf{90.47} \\
                           \cline{2-12}
    & \multirow{2}{*}{PSN} & \multirow{2}{*}{4} & \multirow{2}{*}{Direct} & \multirow{2}{*}{77.37}
                                & APGD & - & - & 69.51 & 44.71 & 78.49 & \textbf{97.25} \\
                      & & & & & SA-PGD & - & - & 69.27 & 43.43 & 78.62 & \textbf{98.56} \\
                            \cline{2-12}
    & \multirow{2}{*}{LIF} & \multirow{2}{*}{8} & \multirow{2}{*}{Direct} & \multirow{2}{*}{78.98}
                                & APGD & 60.69 & 59.30 & 64.79 & 71.92 & 70.06 & \textbf{76.61} \\
                      & & & & & SA-PGD & 61.27 & 60.37 & 64.46 & 73.28 & 70.20 & \textbf{79.74} \\
                            \cline{2-12}
    & \multirow{2}{*}{LIF} & \multirow{2}{*}{4} & \multirow{2}{*}{Poisson} & \multirow{2}{*}{71.81}
                                & APGD & 55.61 & 55.17 & 56.34 & \textbf{61.88} & 59.55 & 60.70 \\
                      & & & & & SA-PGD & 58.59 & 59.37 & 57.14 & \textbf{63.60} & 60.64 & 62.84 \\
  \hline
\multirow{10}{*}{\shortstack{CIFAR \\ -100}}  & \multirow{2}{*}{LIF-2} & \multirow{2}{*}{4} & \multirow{2}{*}{Direct} & \multirow{2}{*}{50.95}
                                & APGD & 81.88 & 84.37 & 83.15 & 87.02 & 86.83 & \textbf{88.28} \\
                      & & & & & SA-PGD & 81.99 & 84.00 & 83.02 & 87.50 & 87.38 & \textbf{89.60} \\
                            \cline{2-12}
    & \multirow{2}{*}{IF} & \multirow{2}{*}{4} & \multirow{2}{*}{Direct} & \multirow{2}{*}{48.27}
                                & APGD & 83.82 & 86.43 & 84.70 & 77.22 & 87.95 & \textbf{89.87} \\
                      & & & & & SA-PGD & 83.79 & 86.25 & 84.19 & 79.49 & 88.75 & \textbf{91.16} \\
                           \cline{2-12}
    & \multirow{2}{*}{PSN} & \multirow{2}{*}{4} & \multirow{2}{*}{Direct} & \multirow{2}{*}{48.53}
                                & APGD & - & - & 84.91 & 78.61 & 90.36 & \textbf{97.90} \\
                      & & & & & SA-PGD & - & - & 84.76 & 78.95 & 90.61 & \textbf{98.82} \\
                            \cline{2-12}
    & \multirow{2}{*}{LIF} & \multirow{2}{*}{8} & \multirow{2}{*}{Direct} & \multirow{2}{*}{51.87}
                                & APGD & 79.21 & 79.76 & 81.43 & 86.46 & 85.07 & \textbf{87.76} \\
                      & & & & & SA-PGD & 79.54 & 80.09 & 80.97 & 87.00 & 85.19 & \textbf{89.14} \\
                            \cline{2-12}
    & \multirow{2}{*}{LIF} & \multirow{2}{*}{4} & \multirow{2}{*}{Poisson} & \multirow{2}{*}{44.97}
                                & APGD & 75.45 & 75.72 & 77.09 & \textbf{80.64} & 80.20 & 80.50 \\
                      & & & & & SA-PGD & 76.58 & 77.48 & 77.34 & \textbf{81.15} & 80.73 & 81.01 \\
  \hline
  \end{tabular}
\end{table*}

% \begin{figure}[t]
%   \centering
%    \includegraphics[width=0.96\linewidth]{ASSG-abla.png}
%    \caption{ASR for attacks using Atan surrogate gradients with different values of $\alpha$ and ASSG with different values of $A$.}
%    \label{assg-abla}
% \end{figure}

\subsection{Ablation Study}

\textbf{Ablation Study on Gradient Approximation.} 
ASSG uses $\omega$ as a parameter to adaptively compute the sharpness parameter $\alpha_{i, t}^l$ with a certain gradient-vanishing degree, based on the distribution of $\left|u_{i, t}^l\right|$ across spatial-temporal dimensions during adversarial attacks. To verify the effectiveness of this input-adaptive and spatial-temporally heterogeneous $\alpha_{i, t}^l$, we select five representative surrogate function forms—Atan, Gaussian (Gauss), Sigmoid (Sig), Rectangular (Rect), and Triangular (Tri)—and apply ASSG to perform adversarial attacks on AT-trained SNNs.

For comparison, we determine a non-adaptive fixed $\alpha$ for each surrogate gradient using ternary search. \cref{ASSG-abla} reports the attack results of SA-PGD. ASSG with different surrogate gradients achieves higher ASR than both the fixed-$\alpha$ baselines and the SOTA methods in \cref{tab1}. This demonstrates that the input-dependent, spatial-temporally adaptive adjustment of surrogate gradient sharpness in ASSG plays a crucial role in accurately estimating input gradients, and that ASSG generalizes well across surrogate gradients sharing a unified form. We also shows the ASR for attacks using Atan surrogate gradients with different values of fixed $\alpha$ and ASSG with different $A$ in Appendix \cref{S42}.

\textbf{Ablation Study on Adversarial Attack.} We use ASSG with Atan as the surrogate gradient and compare the ASR of multiple attack methods across different numbers of iterations. The update rule in SA-PGD is similar to Adam~\cite{kingma2014adam}—incorporating adaptive step sizes and momentum—but it additionally applies per-step $L_\infty$-norm clipping to the update (see \cref{18}). To isolate the effect of this clipping, we construct a baseline attack that uses Adam-style optimization for perturbation updates (see Appendix \cref{S32} for details); we denote this variant Adam-PGD. Moreover, we include the commonly used attacks MI-FGSM~\cite{dong2018boosting}, PGD, and APGD as baselines. Each attack is evaluated with multiple iteration numbers to ensure a rigorous comparison of convergence behavior and final ASR.

\begin{figure}[t]
  \centering
   \includegraphics[width=0.96\linewidth]{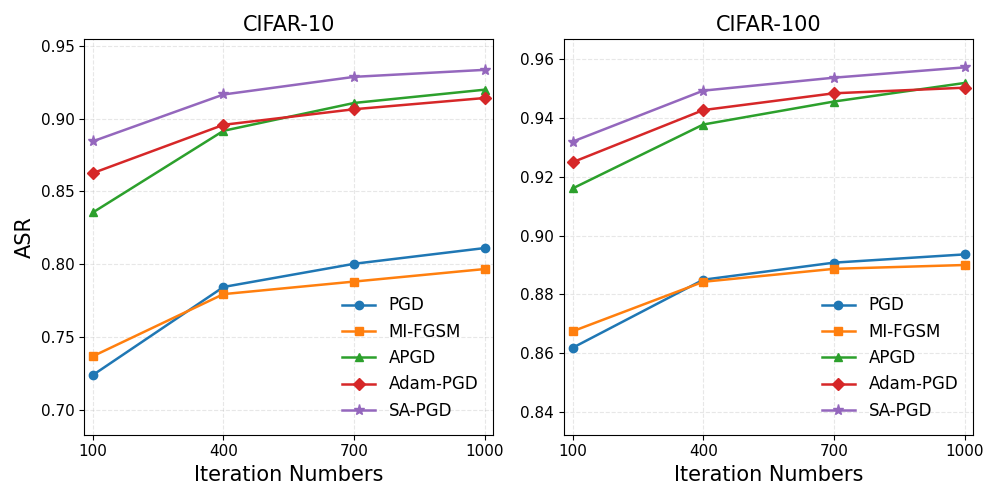}
   \caption{ASR of different attack methods at varying iteration numbers when using ASSG as the surrogate gradient.}
   \label{sapgd-abla}
\end{figure}

As shown in \cref{sapgd-abla}, the ASR continues to increase slightly for all attack methods even after 1000 iterations. This observation demonstrate the stability of the update dynamics described in \cref{13}–\cref{15}. Regardless of the optimization algorithm used, ASSG remains stable throughout the iterations and does not collapse into degraded gradient estimates as the attack iterations increase. PGD and MI-FGSM show notably weaker performance. Adam-PGD, benefiting from its adaptive step-size mechanism, converges faster and achieves higher ASR than APGD at low iteration counts; however, as iterations increase, its ASR drops below APGD. This is because Adam-PGD, lacking per-step $L_\infty$-norm clipping, may produce updates that deviate excessively from the constrained region, causing instability in later stages. In contrast, SA-PGD, which explicitly incorporates the $L_\infty$-norm constraint in \cref{18}, achieves the highest ASR across all iteration numbers, clearly demonstrating its superiority.

\subsection{Generalizability across Configurations}

We also evaluate the generalizability of the proposed method across different configurations, including varying neuron models, network architectures, time steps and encoder. All models in this section are trained by AT.

In the Appendix \cref{S43}, we provide attack results on SNNs with different network architectures and depths. Across all network architectures and depths, both ASSG and SA-PGD effectively improve the ASR, demonstrating the excellent applicability of our method to different network structures.

Furthermore, for SNNs based on the SEWResNet19 architecture, we comprehensively evaluate the generalizability of our proposed method in \cref{tab3}. Since Poisson encoder is a stochastic input transformation, we use Expectation over Transformation (EOT) to compute input gradients during the attack~\cite{athalye2018synthesizing}, with the number of attack iterations set to 50. To reduce the estimation bias in EOT, the update of the surrogate gradient sharpness in ASSG is frozen within the EOT loop. In addition, the PSN model differs significantly from conventional neuron models—it has no fixed threshold or reset mechanism—making it incompatible with BPTR and RGA methods. 

As shown in \cref{tab3}, our method performs well across all configurations. ASSG achieves superior attack effectiveness on SNNs with various neuron models and time steps, with its ASR only slightly lower than HART under Poisson encoder. Notably, ASSG attains over 98\% ASR on the PSN model, substantially outperforming other gradient approximation methods. In contrast, HART shows significantly weaker attack performance on PSN, highlighting the limitations of gradient approximation methods derived from specific neuron dynamics. In comparison, ASSG computes the adaptive surrogate gradient solely based on the input distribution of the function $H$, independent of neuron dynamics, thereby exhibiting much stronger generalizability.
\section{Conclusion}

In this work, we theoretically analyze the gradient vanishing behavior of surrogate functions in SNNs and propose ASSG, an effective gradient approximation method for adversarial attack optimization. Together with the optimization algorithm SA-PGD, it forms a more reliable framework for evaluating the adversarial robustness of SNNs. Extensive experiments demonstrate that our approach consistently outperforms prior methods in both performance and generalizability, advancing the understanding and assessment of robust SNNs.

Moreover, our results reveal that existing studies have overestimated the adversarial robustness of SNNs and underestimated the number of attack iterations required for rigorous robustness evaluation. The results for PSN in \cref{tab3} indicate that, without ASSG, SNNs may even exhibit false robustness. 

In current SNN adversarial training, the inner maximization in \cref{2} is typically approximated using 5–10 steps of PGD with the same surrogate gradients used during training (i.e., STBP). Our findings show that this approximation is far from optimal, severely compromising the effectiveness of adversarial training. Hence, there is a pressing need to rethink how to design efficient and reliable methods for solving the inner maximization problem in SNN adversarial training.
{
    \small
    \bibliographystyle{ieeenat_fullname}
    \bibliography{main}
}

\clearpage
\setcounter{page}{1}

% ------- Section S1, S2, ... -------
\setcounter{section}{0}
\renewcommand{\thesection}{S\arabic{section}}
\renewcommand{\thesubsection}{S\arabic{section}.\arabic{subsection}}

% ★★★ Hyperref anchor fix ★★★
\renewcommand{\theHsection}{S.\arabic{section}}
\renewcommand{\theHsubsection}{S.\arabic{section}.\arabic{subsection}}

%=============== FIGURE ===============
\setcounter{figure}{0}
\renewcommand{\thefigure}{S\arabic{figure}}
\renewcommand{\theHfigure}{S.\arabic{figure}}

%=============== TABLE ===============
\setcounter{table}{0}
\renewcommand{\thetable}{S\arabic{table}}
\renewcommand{\theHtable}{S.\arabic{table}}

%=============== EQUATION ===============
\setcounter{equation}{0}
\renewcommand{\theequation}{S\arabic{equation}}
\renewcommand{\theHequation}{S.\arabic{equation}}

\maketitlesupplementary

\textbf{Overview. }The structure of this appendix is as follows:

\textbf{\cref{S1}} provides the proofs of the main theoretical results presented in this paper.

\textbf{\cref{S2}} describes the detailed dynamics of the neuron models used in our experiments.

\textbf{\cref{S3}} presents the full parameter settings for adversarial training and adversarial attacks.

\textbf{\cref{S4}} includes additional experimental results, including: 
\begin{itemize}
    \item Extended SA-PGD loss curves with more iterations.
    \item Further ablation studies on more training methods.
    \item Computational overhead comparisons for different gradient approximation methods.
    \item Experiments analyzing the effect of the relaxation term.
\end{itemize}

\section{Proofs and Derivations}
\label{S1}

\begin{TheoremS}
For any surrogate function $g(x)$, the gradient-vanishing degree function $G(x) = \int_{-x}^{x} g(t) \, dt,x \geq 0$ is a concave function on $[0,+\infty)$
\label{st1}
\end{TheoremS}

\begin{proof}
Since $g$ is even, we have $g(-t)=g(t)$ for all $t\in\mathbb{R}$, and hence
\begin{equation}\label{eq:G-even}
G(x) = \int_{-x}^{x} g(t)\,dt
     = 2\int_{0}^{x} g(t)\,dt, 
     \quad x \ge 0.
\end{equation}

We first express the slope of the secant line of $G$ over an interval $[x,y]$ with $0 \le x < y$ in terms of the average of $g$ on $[x,y]$. Using \cref{eq:G-even}, we obtain
\begin{align*}
\frac{G(y)-G(x)}{y-x}
&= \frac{2\int_{0}^{y} g(t)\,dt - 2\int_{0}^{x} g(t)\,dt}{y-x} \\
&= \frac{2}{y-x}\int_{x}^{y} g(t)\,dt.
\end{align*}

For any $0\le x < y$ and any $t\in[x,y]$, as $g$ is non-increasing on $[0,+\infty)$, we have:
\[
g(y) \le g(t) \le g(x).
\]
Integrating this inequality over $t\in[x,y]$ and dividing by $y-x>0$, we obtain
\[
g(y) \le \frac{1}{y-x}\int_{x}^{y} g(t)\,dt \le g(x).
\]
This yields
\begin{equation}\label{eq:slope-bounds}
2g(y) \le \frac{G(y)-G(x)}{y-x} \le 2g(x),
\quad 0\le x<y.
\end{equation}

We now compare slopes on adjacent intervals. Let $0 \le x < y < z$. Applying \cref{eq:slope-bounds} to the pair $(x,y)$ gives
\[
\frac{G(y)-G(x)}{y-x} \;\ge\; 2g(y),
\]
while applying \cref{eq:slope-bounds} to the pair $(y,z)$ gives
\[
\frac{G(z)-G(y)}{z-y} \;\le\; 2g(y).
\]
Combining these two inequalities, we obtain
\begin{equation}\label{eq:secant-monotone}
\frac{G(y)-G(x)}{y-x} \ge \frac{G(z)-G(y)}{z-y},
\quad 0\le x<y<z.
\end{equation}

The inequality \cref{eq:secant-monotone} states that the secant slopes of $G$ are nonincreasing as the interval moves to the right. This property is equivalent to concavity. For completeness, we sketch the standard argument.

Fix $0 \le x < z$ and $\lambda\in(0,1)$, and define
\[
y = \lambda z + (1-\lambda)x,
\]
so that $x<y<z$ and
\[
y-x = \lambda(z-x), \quad z-y = (1-\lambda)(z-x).
\]
Substituting into \cref{eq:secant-monotone} and using $z-x>0$ yields
\[
\frac{G(y)-G(x)}{\lambda(z-x)} 
\;\ge\; 
\frac{G(z)-G(y)}{(1-\lambda)(z-x)}.
\]
Multiplying both sides by $\lambda(1-\lambda)(z-x) > 0$ gives
\[
(1-\lambda)\bigl(G(y)-G(x)\bigr) \;\ge\; \lambda\bigl(G(z)-G(y)\bigr).
\]
Rearranging terms, we obtain
\[
(1-\lambda)G(y) + \lambda G(y) \;\ge\; \lambda G(z) + (1-\lambda)G(x),
\]
that is,
\[
G(y) \;\ge\; \lambda G(z) + (1-\lambda)G(x)
= \lambda G\bigl(z\bigr) + (1-\lambda)G\bigl(x\bigr).
\]
Recalling that $y = \lambda z + (1-\lambda)x$, this is precisely the concavity inequality
\[
G\bigl(\lambda z + (1-\lambda)x\bigr)
\;\ge\;
\lambda G(z) + (1-\lambda)G(x)
\]

Hence $G$ is concave on $[0,+\infty)$.
\end{proof}

\begin{CorollaryS}
    \label{sc1}
    Let $g(x)=\alpha k(\alpha x)$ a surrogate function, and the corresponding  gradient-vanishing degree function $\int_{-x}^{x} \alpha k(\alpha t) \, dt =  \int_{-\alpha x}^{\alpha x} k(\alpha t) \, d\alpha t = G(\alpha x), x \geq 0$. Assume that $x$ follows a probability distribution $p(x)$ with finite expectation $\mathbb{E}[x] < +\infty$. Then, for any constant $A \in [0,1]$, the expected gradient-vanishing degree satisfies 
    \begin{equation}
    \mathbb{E}_{x\!\sim\!p(x)}[G(\alpha x)] \le A
    \;\text{if}\;
    \alpha \le \frac{G^{-1}(A)}{\mathbb{E}[x]}
    \label{s9}
    \end{equation}
\end{CorollaryS}

\begin{proof}
Recall that for $x\ge 0$,
\[
G(\alpha x)=\int_{-x}^{x}\alpha k(\alpha t)\,dt=\int_{-\alpha x}^{\alpha x}k(u)\,du ,
\]
hence $G:[0,+\infty)\to[0,1]$ is nondecreasing and satisfies $G(0)=0$ (and typically $\lim_{z\to\infty}G(z)=1$). According to Theorem \ref{st1}, $G$ is concave on $[0,+\infty)$.

Let $X\sim p(x)$ be a nonnegative random variable with $\mathbb{E}[X]<+\infty$.
Since $G$ is concave, Jensen's inequality gives
\[
\mathbb{E}\bigl[G(\alpha X)\bigr]\le G\bigl(\mathbb{E}[\alpha X]\bigr)=G\bigl(\alpha\,\mathbb{E}[X]\bigr).
\]
Therefore, for any $A\in[0,1]$, if
\[
G\bigl(\alpha\,\mathbb{E}[X]\bigr)\le A,
\]
then automatically $\mathbb{E}[G(\alpha X)]\le A$.

Because $G$ is nondecreasing on $[0,+\infty)$, the condition $G(\alpha\mathbb{E}[X])\le A$ is equivalent to
\[
\alpha\,\mathbb{E}[X]\le G^{-1}(A),
\]
where $G^{-1}(A)$ denotes the (generalized) inverse
\[
G^{-1}(A):=\inf\{z\ge 0:\;G(z)\ge A\}.
\]
Thus, it suffices that
\[
\alpha \le \frac{G^{-1}(A)}{\mathbb{E}[X]}.
\]
Combining with the Jensen bound yields
\[
\mathbb{E}_{x\sim p(x)}[G(\alpha x)]\le A
\quad\text{if}\quad
\alpha \le \frac{G^{-1}(A)}{\mathbb{E}[x]},
\]
which proves the corollary.
\end{proof}

According to Corollary \ref{sc1}, we present below a series of specific surrogate functions along with their corresponding methods for computing the adaptive sharpness.

\paragraph{(1) Arctangent (Atan) surrogate}
\begin{equation}
g_{\mathrm{Atan}}(x) = \frac{\alpha}{2 \left[1 + (\frac{\pi}{2} \alpha x)^2 \right]}
\label{Atan}
\end{equation}
\begin{equation}
    \alpha_{i,t}^l = \frac{2}{\pi\,\mathbb{E}[\left\lvert u_{i,t}^l \right\rvert]}\tan\!\left(\frac{\pi}{2}A\right)
    \label{AS-A}
\end{equation}

\paragraph{(2) Gaussian (Gauss) surrogate}
\begin{equation}
g_{\mathrm{Gauss}}(x)
= \frac{\alpha}{\sqrt{2\pi}}\exp\!\left(-\frac{(\alpha x)^2}{2}\right)
\label{G}
\end{equation}
\begin{equation}
    \alpha_{i,t}^l = \Phi ^{-1}(\frac{1+A}{2})\frac{1}{\mathbb{E}\left[\left|u_{i, t}^l\right|\right]}
    \label{AS-G}
\end{equation}
where $\Phi$ denotes standard Gaussian PDF.

\paragraph{(3) Sigmoid (Sig) surrogate}
\begin{equation}
g_{\mathrm{Sig}}(x)
= \alpha\,\sigma(\alpha x)\bigl(1-\sigma(\alpha x)\bigr)
\label{S}
\end{equation}
where $\sigma(u)=\frac{1}{1+e^{-u}}$.
\begin{equation}
    \alpha_{i,t}^l = \ln(\frac{2}{1-A}-1)\frac{1}{\mathbb{E}\left[\left|u_{i, t}^l\right|\right]}
    \label{AS-S}
\end{equation}

\paragraph{(4) Rectangular (Rect) surrogate}
\begin{equation}
g_{\mathrm{Rect}}(x)
= \frac{\alpha}{2}\,\mathbf{1}\!\left\{|x|\le \frac{1}{\alpha}\right\}
\label{Rect}
\end{equation}
\begin{equation}
    \alpha_{i,t}^l = A\frac{1}{\mathbb{E}\left[\left|u_{i, t}^l\right|\right]}
    \label{AS-R}
\end{equation}

\paragraph{(5) Triangular (Tri) surrogate}
\begin{equation}
g_{\mathrm{Tri}}(x)
= \alpha\left(1-\alpha|x|\right)\,
\mathbf{1}\!\left\{|x|\le \frac{1}{\alpha}\right\}
\label{Tri}
\end{equation}
\begin{equation}
    \alpha_{i,t}^l = (1-\sqrt{1-A})\frac{1}{\mathbb{E}\left[\left|u_{i, t}^l\right|\right]}
    \label{AS-T}
\end{equation}

It can be observed that the adaptive sharpness parameters of different surrogate gradients share a unified form: $\alpha_{i, t}^l=\frac{\omega}{\mathbb{E}\left[\left|u_{i, t}^l\right|\right]}$. For the Atan, Gauss, and Sig functions, the valid range of $\omega$ is $[0,+\infty)$. In contrast, for the Rect and Tri functions, the valid range of $\omega$ is $[0,1]$. This is because for $g_{\mathrm{Rect}}$ and $g_{\mathrm{Tri}}$, the degree of gradient vanishing reaches its upper bound of 1 when $|x| > \frac{1}{\alpha}$, which leads to the constraint $\mathbb{E}\left[\left|u_{i, t}^l\right|\right]\le \frac{1}{\alpha}$ when deriving the inverse function. However, this prerequisite cannot be guaranteed for all inputs and neurons during adversarial attacks. Therefore, we unify the range of $\omega$ of all surrogate functions to $[0,+\infty)$.

\section{Spiking Neuron Models}
\label{S2}

\textbf{LIF-2.} We use LIF-2 to denote a variant of the LIF neuron model, which has been adopted in prior works such as~\cite{bu2023rate,hao2023threaten,lun2025towards,ding2022snn,liu2024enhancing,ding2024enhancing}. The dynamics of LIF-2 are given as follows:
\begin{equation}
    \boldsymbol{V}^l(t+1) = \lambda R\!\left(\boldsymbol{V}^l(t), \boldsymbol{s}^l(t)\right) + \boldsymbol{W}^l \boldsymbol{s}^{l-1}(t+1),
    \label{lif2f}
\end{equation}
\begin{equation}
R\!\left(\boldsymbol{V}^l(t), \boldsymbol{s}^l(t)\right)= 
\begin{cases}
0, & \boldsymbol{s}^l(t) = 1,\\[3pt]
\boldsymbol{V}^l(t), & \boldsymbol{s}^l(t) = 0.
\end{cases}
\label{lif2r}
\end{equation}
The main difference between LIF-2 and the standard LIF neuron is that LIF-2 does not apply the $1-\lambda$ normalization factor to the input current.

\textbf{IF.} The IF neuron further simplifies this by setting the decay coefficient to 1:
\begin{equation}
    \boldsymbol{V}^l(t+1) = R\!\left(\boldsymbol{V}^l(t), \boldsymbol{s}^l(t)\right) + \boldsymbol{W}^l \boldsymbol{s}^{l-1}(t+1),
    \label{if2f}
\end{equation}
\begin{equation}
R\!\left(\boldsymbol{V}^l(t), \boldsymbol{s}^l(t)\right)= 
\begin{cases}
0, & \boldsymbol{s}^l(t) = 1,\\[3pt]
\boldsymbol{V}^l(t), & \boldsymbol{s}^l(t) = 0.
\end{cases}
\label{if2r}
\end{equation}

\textbf{PSN.} The PSN is a neuron model that enables parallel processing along the temporal dimension, and its dynamics differ fundamentally from those of conventional neurons~\cite{fang2023parallel}. PSN removes the reset mechanism and treats the firing threshold at each time step as a learnable parameter:
\begin{equation}
\boldsymbol{V}^l = \boldsymbol{W}^l_T \boldsymbol{X}^l,\boldsymbol{W}^l_T \in \mathbb{R}^{T \times T}, \boldsymbol{X}^l \in \mathbb{R}^{T \times N}
\label{psn1}
\end{equation}
\begin{equation}
\boldsymbol{S}^l = H(\boldsymbol{V}^l-\boldsymbol{B}^l), \boldsymbol{B}^l \in \mathbb{R}^T, \boldsymbol{S}^l \in\{0,1\}^{T \times N}
\label{psn2}
\end{equation}
where $\boldsymbol{X}^l$, $\boldsymbol{V}^l$ and $\boldsymbol{S}^l$ denote the input current, membrane potential, and spike sequence, respectively, and $\boldsymbol{W}^l_T$ and $\boldsymbol{B}^l$ represent the learnable temporal weight matrix and threshold parameters. This formulation allows the membrane potential $\boldsymbol{V}^l(t)$ at each time step to integrate information from all temporal positions, rather than relying solely on the previous step as in conventional neuron models.

\section{Detailed Experimental Settings}
\label{S3}

\subsection{Adversarial Training Settings}
\label{S31}

In this work, we train robust SNN models using commonly adopted adversarial training schemes for evaluation. Both standard adversarial training (AT) and TRADES generate adversarial examples using PGD with a perturbation budget of $\varepsilon=8/255$, a step size of $\eta=4/255$, and 5 attack iterations. RAT and AT+SR follow the same AT setup but additionally incorporate their respective regularization terms, with the regularization coefficient set to 0.004 for both methods. 

We adopt Temporal Efficient Training (TET)~\cite{duan2022temporal} as the loss function and optimize the models using SGD with a momentum of 0.9. All models are trained for 100 epochs. For SNNs using neuron models other than PSN, we employ a triangular surrogate gradient during training:
\begin{equation}
\frac{dH}{dx} \approx 2 \cdot \max \left\{0, 1-\left|x\right|\right\}
\end{equation}
PSN-based SNNs are trained using the Atan surrogate gradient with $\alpha = 4$.
\begin{figure*}[t]
  \centering
   \includegraphics[width=0.96\linewidth]{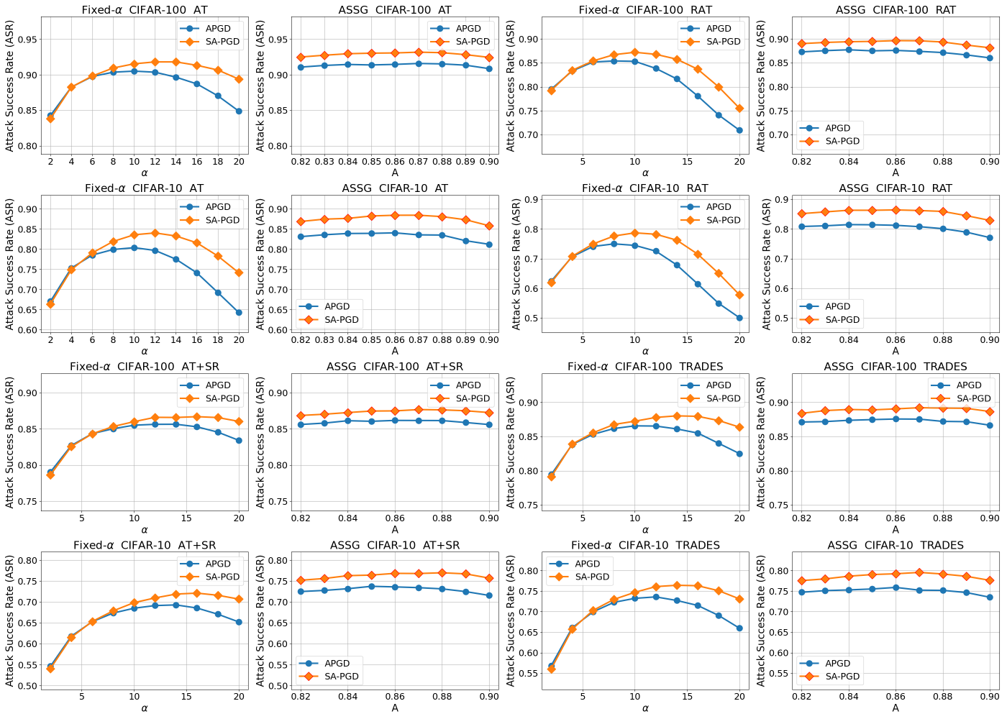}
   \caption{ASR for attacks using Atan surrogate gradients with different values of $\alpha$ and ASSG with different values of $A$.}
   \label{sassg}
\end{figure*}

\begin{table*}[h]
  \centering
  \caption{Computational cost of training one epoch on CIFAR-10}
  \label{tabs1}
  \begin{tabular}{ccccccc}
  \hline
  & BPTR & RGA & PDSG & HART & STBP & ASSG \\
   \hline
  Runtime(s) & 7.7 & 8.0 & 13.7 & 11.2 & 11.3 & 15.4 \\
  \hline
  Memory Usage(MB) & 7161 & 8171 & 11265 & 9473 & 9473 & 13577 \\
  \hline
  \end{tabular}
\end{table*}

\subsection{Adversarial Attack Settings}
\label{S32}

For RGA and PDSG, we follow the parameter settings provided in their original papers~\cite{bu2023rate,lun2025towards}. For ASSG, we set $\beta_1 = 0.9$ and $\beta_2 = 0.9$ when Poisson encoding is not used.
When Poisson encoding is applied, we instead use $\beta_1 = 0.99$ and $\beta_2 = 0.99$. The initial values of $M_{i,t}^l$ and $D_{i,t}^l$ are set to $M_{0}=1.0$ and $D_{0}=0.0$, respectively, and the relaxation coefficient $\gamma = 1.5$. We search for the optimal value of $A$ within the range $[0.82,0.9]$ with a step size of 0.01. Although the ASR values of ASSG reported in \cref{tab1}-\cref{tab3} correspond to the best-performing $A$, \cref{assg-abla} and \cref{sassg} demonstrate that the attack performance of ASSG is robust to the choice of $A$. 

For SA-PGD, we set $\beta_m=0.5$ and $\beta_v=0.8$. When attacking SNNs with Poisson encoding, we estimate the gradient using the Expectation-over-Transformation (EOT) technique~\cite{athalye2018synthesizing}, with 10 EOT samples per iteration.

We design Adam-PGD as a baseline for comparison with SA-PGD. The update rule in Adam-PGD follows the Adam optimization mechanism, with an additional projection step to ensure that the updated input remains within the $L_\infty$ constraint:
\begin{equation}
\boldsymbol{m}_{k} = \beta_m \boldsymbol{m}_{k-1} + (1-\beta_m) \nabla_{\boldsymbol{x}} \mathcal{L}\left(\boldsymbol{f}_{\boldsymbol{\theta}}\left(\boldsymbol{x}_{k}\right), y\right)
\label{sadam1}
\end{equation}
\begin{equation}
\boldsymbol{v}_{k} = \beta_v \boldsymbol{v}_{k-1} + (1-\beta_v)\nabla_{\boldsymbol{x}} \mathcal{L}\left(\boldsymbol{f}_{\boldsymbol{\theta}}\left(\boldsymbol{x}_{k}\right), y\right)^2
\label{sadam2}
\end{equation}
\begin{equation}
\widehat{\boldsymbol{m}}_k = \frac{\boldsymbol{m}_k}{1 - \beta_m^k}, 
\qquad
\widehat{\boldsymbol{v}}_k = \frac{\boldsymbol{v}_k}{1 - \beta_v^k},
\label{adam-bias}
\end{equation}
\begin{equation}
\boldsymbol{x}_{k+1}
= \Pi_{\varepsilon}^{\infty} \left(\boldsymbol{x}_k
+ \eta_k \frac{\widehat{\boldsymbol{m}}_k}{\sqrt{\widehat{\boldsymbol{v}}_k} + \xi}\right)
\label{adam-update}
\end{equation}
where $\eta_k$ adopt the same learning rate adjustment strategy as in APGD and SA-PGD. We set $\beta_m=0.8$ and $\beta_v=0.9$ in Adam-PGD.

\section{Additional Results}
\label{S4}

\begin{table*}[h]
  \centering
  \caption{ASR on different network architectures and depths.}
  \label{tab2}
  \begin{tabular}{c|c|cccccccc}
  \hline
 Datasets & Architectures & Clean & Attack & BPTR & RGA & PDSG & HART & STBP & \textbf{ASSG} \\
   \hline
\multirow{10}{*}{\shortstack{CIFAR \\ -10}} & \multirow{2}{*}{VGG9} & \multirow{2}{*}{82.09}
                            & APGD & 62.86 & 67.15 & 67.60 & 77.39 & 72.37 & \textbf{83.97} \\
                      & & & SA-PGD & 62.86 & 67.59 & 67.04 & 79.86 & 72.21 & \textbf{88.16} \\
                            \cline{2-10}
    & \multirow{2}{*}{SEWResNet19} & \multirow{2}{*}{78.69} & APGD & 66.45 & 63.94 & 69.31 & 77.22 & 75.38 & \textbf{84.06} \\
                                                      & & & SA-PGD & 66.85 & 65.08 & 68.76 & 79.49 & 75.11 & \textbf{88.44} \\
                           \cline{2-10}
    & \multirow{2}{*}{SEWResNet31} & \multirow{2}{*}{77.50} & APGD & 65.72 & 60.82 & 68.38 & 75.73 & 74.49 & \textbf{83.18} \\
                                                      & & & SA-PGD & 65.72 & 61.63 & 67.27 & 77.90 & 74.31 & \textbf{87.38} \\
                            \cline{2-10}
    & \multirow{2}{*}{SEWResNet43} & \multirow{2}{*}{74.80} & APGD & 57.86 & 59.25 & 66.15 & 75.61 & 72.79 & \textbf{88.05} \\
                                                      & & & SA-PGD & 56.87 & 59.08 & 64.73 & 77.41 & 71.76 & \textbf{91.43} \\
                            \cline{2-10}
  \hline
\multirow{10}{*}{\shortstack{CIFAR \\ -100}} & \multirow{2}{*}{VGG9} & \multirow{2}{*}{54.74} 
                            & APGD & 77.28 & 82.16 & 81.10 & 88.06 & 84.26 & \textbf{89.94} \\
                      & & & SA-PGD & 77.33 & 82.16 & 80.63 & 88.94 & 84.28 & \textbf{91.77} \\
                            \cline{2-10}
    & \multirow{2}{*}{SEWResNet19} & \multirow{2}{*}{51.90} & APGD & 82.83 & 82.50 & 84.21 & 89.88 & 88.22 & \textbf{91.60} \\
                                                      & & & SA-PGD & 82.99 & 83.07 & 84.00 & 90.80 & 88.26 & \textbf{93.19} \\
                           \cline{2-10}
    & \multirow{2}{*}{SEWResNet31} & \multirow{2}{*}{51.02} & APGD & 82.38 & 80.49 & 83.86 & 88.78 & 87.60 & \textbf{89.74} \\
                                                      & & & SA-PGD & 82.68 & 81.08 & 83.59 & 89.59 & 87.66 & \textbf{91.44} \\
                            \cline{2-10}
    & \multirow{2}{*}{SEWResNet43} & \multirow{2}{*}{49.27} & APGD & 83.24 & 79.37 & 84.72 & 88.00 & 87.66 & \textbf{89.00} \\
                                                      & & & SA-PGD & 83.29 & 80.22 & 84.46 & 88.69 & 87.82 & \textbf{90.60} \\
                            \cline{2-10}
    \hline     
  \end{tabular}
\end{table*}

\begin{figure}[h]
  \centering
   \includegraphics[width=0.96\linewidth]{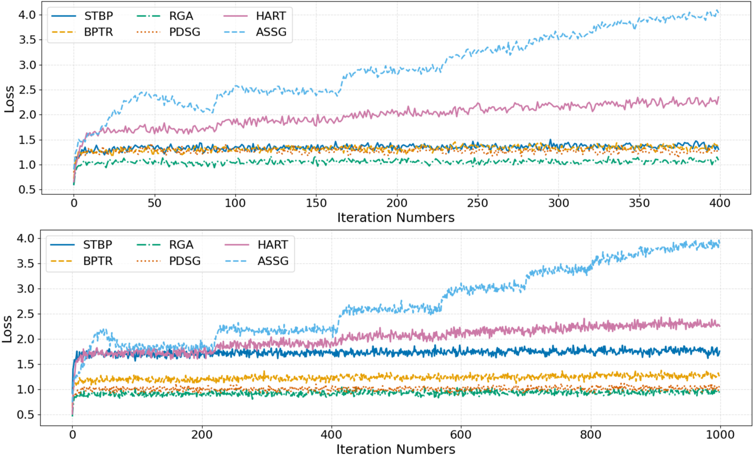}
   \caption{The loss curves of SA-PGD for different gradient approximation methods and iteration numbers.}
   \label{sloss}
\end{figure}

\begin{table*}[t]
  \centering
  \caption{Comparison of ASR between ASSG with and without the relaxation term.}
  \label{tabs2}
  \begin{tabular}{c|c|cccccccc}
  \hline
 Datasets & Methods & $\gamma=0$ & $\gamma=0.5$ & $\gamma=1.0$ & $\gamma=1.5$ & $\gamma=2.0$ & $\gamma=2.5$ & $\gamma=3.0$ & $\gamma=3.5$  \\
   \hline
\multirow{4}{*}{\shortstack{CIFAR \\ -10}} & AT & 86.33 & 87.45 & 87.97 & 88.44 & 88.83 & 88.96 & 89.16 & 89.30 \\
                                          & RAT & 85.39 & 85.92 & 86.32 & 86.44 & 86.78 & 86.95 & 87.22 & 87.27 \\
                                        & AT+SR & 75.08 & 76.13 & 76.87 & 77.00 & 77.48 & 77.68 & 77.75 & 77.89 \\
                                       & TRADES & 77.00 & 78.36 & 78.97 & 79.59 & 79.86 & 80.19 & 80.49 & 80.57 \\
  \hline
\multirow{4}{*}{\shortstack{CIFAR \\ -100}} & AT & 92.45 & 92.87 & 93.14 & 93.19 & 93.40 & 93.58 & 93.67 & 93.76 \\
                                          & RAT & 89.02 & 89.44 & 89.55 & 89.64 & 89.75 & 89.80 & 89.93 & 89.89 \\
                                        & AT+SR & 86.72 & 87.22 & 87.53 & 87.68 & 87.87 & 87.92 & 87.98 & 88.11 \\
                                       & TRADES & 88.43 & 88.91 & 89.05 & 89.22 & 89.35 & 89.40 & 89.46 & 89.43 \\
    \hline              
  \end{tabular}
\end{table*}

\subsection{Loss Curves of more Iteration Numbers}
\label{S41}

We further plot the loss curves of different gradient approximation methods under SA-PGD at 400 and 1000 attack iterations to more comprehensively examine their convergence behavior. As shown in \cref{sloss}, ASSG only begins to converge when the iteration reaches 1000 steps, demonstrating its ability to continuously and effectively optimize adversarial perturbations. This requirement for extremely large iteration budgets also highlights the inherent difficulty of generating adversarial examples for SNNs.

\subsection{Additional Comparative Experiments between Fixed-$\alpha $ and ASSG}
\label{S42}

In this section, we present additional results on models trained with various adversarial training methods, reporting the ASR for attacks using Atan surrogate gradients with different values of fixed-$\alpha$ and ASSG with different values of $A$ in \cref{Atan}-\cref{AS-A}. As shown in the \cref{sassg}, the results across more models further demonstrate that ASSG's input-dependent, spatio-temporal adaptive adjustment of the sharpness parameter $\alpha_{i,t}^l$ is superior to directly searching for a globally optimal fixed sharpness parameter $\alpha$.

\subsection{Results on Different Network Architectures and Depths}
\label{S43}

\cref{tab2} reports the attack success rates of various attack methods on SNNs with different network architectures and depths. All target models are trained with AT, and the training settings are consistent with \cref{S31}. ASSG demonstrates significantly stronger attack performance across different network architectures and depths, indicating its strong generalization capability with respect to network structures.

\subsection{Computational Cost Comparison}
\label{S44}

All adversarial attacks are implemented and evaluated in PyTorch on a single NVIDIA A100-PCIE-40GB GPU. We compare the memory usage and runtime of different surrogate gradient methods under the same settings. Specifically, we report the GPU memory consumption and time required to execute 100 steps of SA-PGD on a single CIFAR-10 batch (batch size = 256). As shown in the \cref{tabs1}, although ASSG incurs additional memory and runtime due to its spatio-temporally heterogeneous sharpness parameters $\alpha_{i,t}^l$, this overhead remains acceptable given the substantial improvements in attack effectiveness.

\subsection{Effectiveness of the Relaxation Term}
\label{S45}

ASSG incorporates a relaxation term that leverages second-order moment information to prevent underestimation of $\mathbb{E}[\lvert u_{i,t}^l \rvert]$. Without this term, the computed sharpness parameter $\alpha_{i,t}^l$ may become excessively large, causing the gradient-vanishing degree to exceed the theoretical upper bound. \Cref{tabs2} presents a comparison of ASR between ASSG with different relaxation coefficients. As the appropriate range for tuning the parameter $A$ depends on factors such as the relaxation coefficient, the dataset, and the neuron model, we use a ternary search over the interval $[0.6,0.95]$ to identify the optimal $A$ under each configuration.

As shown in \cref{tabs2}, introducing the relaxation term effectively improves the ASR of ASSG. Although the ASR is relatively robust to the choice of the relaxation coefficient, larger relaxation coefficients generally yield higher ASR. We leave a deeper investigation of this phenomenon—particularly the role of second-order moment information in improving gradient estimation—to future work.

\end{document}